\documentclass{ifacconf}
\usepackage{graphicx}
\graphicspath{{figure/}{graphics/plots/}}
\usepackage[format=plain]{caption}
\usepackage{bm}
\usepackage{mleftright}
\usepackage{threeparttable}

\usepackage{natbib}
\usepackage{tikz}
\usepackage{adjustbox}
\usepackage{pgfplots}
\usepgfplotslibrary{groupplots}
\usetikzlibrary{pgfplots.groupplots}
\usetikzlibrary{plotmarks}
\usetikzlibrary{calc}
\usepackage{comment}
\usepgfplotslibrary{external}
\pgfplotsset{compat=newest}

\usetikzlibrary{matrix}
\pgfplotsset{grid style={dashed,gray}}
\usepackage{amsmath} 
\usepackage{amssymb}  
\usepackage{color,soul}
\usepackage{bibrange}
\usepackage{hvline}
\usepackage{url}
\usepackage{multirow}
\usepackage{booktabs}
\usepackage{epstopdf}

\usepackage[ruled,linesnumbered,algo2e]{algorithm2e}
\usepackage{parskip}
\usepackage{hyphenat}
\usepackage{threeparttable}
\usepackage{subcaption}
\usepackage{lpic}
\usepackage{makecell}
\makeatletter
\let\old@ssect\@ssect 
\makeatother

\usepackage[hidelinks]{hyperref}
\makeatletter
\def\@ssect#1#2#3#4#5#6{%
  \NR@gettitle{#6}
  \old@ssect{#1}{#2}{#3}{#4}{#5}{#6}
}
\makeatother
\makeatletter
\usepackage{siunitx}
\usepackage{nicematrix}
\newcommand{\nosemic}{\renewcommand{\@endalgocfline}{\relax}}
\newcommand{\dosemic}{\renewcommand{\@endalgocfline}{\algocf@endline}}

\providecommand{\FullStop}{\text{~\@.\xspace}}
\providecommand{\Comma}{\text{~,\xspace}}
\newcommand{\vect}[1]{\boldsymbol{\mathbf{#1}}}

\providecommand{\kuka}{\textsc{KUKA} LBR iiwa 14 R820\xspace}
\providecommand{\comau}{\textsc{Comau} Racer 7-1.4\xspace}
\providecommand{\panda}{\textsc{Franka Emika} Panda\xspace}

\begin{document}
\begin{frontmatter}

\title{Singularity Avoidance with Application to Online Trajectory Optimization for Serial Manipulators} 
\author[First]{F. Beck\thanksref{footnoteinfo}},
\author[First]{M.N. Vu\thanksref{footnoteinfo}},
\author[First]{C. Hartl-Nesic},
\author[First,Second]{A. Kugi}

\address[First]{Automation and Control Institute (ACIN), 
   TU Wien, Vienna, Austria (e-mail: \{beck, vu, hartl, kugi\}@acin.tuwien.ac.at).}
\address[Second]{Center for Vision, Automation and Control, AIT Austrian Institute of Technology, Vienna, Austria (e-mail: \{andreas.kugi\}@ait.ac.at)}
\thanks[footnoteinfo]{These authors contributed equally to this work.}
\begin{abstract}                
    This work proposes a novel singularity avoidance approach for real-time trajectory optimization based on known singular configurations. 
    The focus of this work lies on analyzing kinematically singular configurations for three robots with different kinematic structures, i.e., the \comau, the \kuka, and the \panda, and exploiting these configurations in form of tailored potential functions for singularity avoidance.   
    Monte Carlo simulations of the proposed method and  the commonly used manipulability maximization approach are performed for comparison. The numerical results show that the average computing time can be reduced and shorter trajectories in both time and path length are obtained with the proposed approach.
\end{abstract}


\end{frontmatter}
\section{Introduction}
\label{section: introduction}

Several important tasks in robotics require compliance in the robot's end-effector including handling tasks, such as the peg-in-hole task, see, e.g., \cite{park2017compliance} and \cite{song2021peg}, or more recently tasks in physical human-robot interaction (pHRI), see, e.g., \cite{sharifi2021impedance} and \cite{li2018stable}. 
To this end, control concepts enabling compliance in the end-effector, e.g., prescribing a specific impedance as done in~\cite{Ott2008}, are required. 
However, such control concepts in Cartesian space rely on the non-singularity of the manipulator Jacobian or of the pseudo-inverse in the redundant case. To ensure that no singularity occurs during control execution, approaches can be divided into two general categories. 

In the first category, additional measures are taken in the controller to ensure the invertibility of the Jacobian during execution even at singular reference configurations. 
This was investigated intensively in the area of numerical inverse kinematics algorithms. Popular approaches include damped least-squares inverse solutions, see, e.g,~\cite{Chiaverini1997, Buss2005}, and singular value filtering, see, e.g.,~\cite{Colome2015}. 
Furthermore, the manipulability measure proposed by \cite{Yoshikawa1985} can be used as a proxy for singularity avoidance. 
The manipulability measure is proportional to the volume of the manipulability ellipsoid of the manipulator. 
Therefore, a value larger than zero of this measure implies the non-singularity of the manipulator Jacobian. 
However, this manipulability measure is not directly a distance measure to singularities since in degenerate cases of the manipulability ellipsoid the volume can be large even if one direction almost collapses. 
Manipulability maximization for inverse kinematics is done, e.g., in~\cite{Dufour2017}.
A potential function on the torque level, as an additive impedance, based on the manipulability measure is proposed in~\cite{Ott2008} for singularity avoidance.
Due to the complexity introduced by maximizing the manipulability measure, an optimization approach using a dynamic neural network is introduced in~\cite{Jin2017} for tracking control including the consideration of joint velocity limits.


In the second category, trajectories are planned such that no singularities occur.
This has the main advantage that other objectives and constraints can be taken into account during planning. Similar to optimization-based inverse kinematics, these aspects can be included over the whole trajectory horizon and not only locally at every point in the inverse kinematics.
Consider for example an obstacle-free trajectory that is executed with a task-space controller implementing one of the singularity avoidance concepts of the first category. 
Since the controller does not know anything about the obstacles in the environment it may happen that due to singularity avoidance the robot deviates from the planned trajectory and crashes into an obstacle. Furthermore, the controller only reacts instantaneously and is unable to predict a potential crash or a violation of the joint limit.
In contrast, if the planned trajectory is obstacle free and singularity free, the trajectory can be executed without violating any of these constraints, assuming the trajectory tracking controller is able to follow the desired trajectory.
Closely related to the inverse kinematics approaches, a joint-space trajectory generation algorithm which maximizes manipulability based on a task-space trajectory is presented in~\cite{Guilamo2006}. 
This algorithm is classified as a search-based method which guarantees resolution completeness and global optimality but is computationally quite complex. 
Furthermore, it does not solve the full planning problem but relies on a known task-space reference and does not provide any possibility to include further optimization criteria.
In~\cite{Menasri2013}, a path planning approach for manipulability maximization and obstacle avoidance using a bi-level genetic algorithm is proposed. The algorithm is demonstrated in simulation on a robot with 5 degrees of freedom (DoF), but no results with respect to execution time and the real-time capabilities are reported.
A fast manipulability maximization for trajectory optimization using Gaussian processes is done in~\cite{Maric2019}. The authors demonstrate fast singularity avoidance combined with obstacle avoidance in planning. However, general joint-space constraints are not considered, which typically increase the computation time significantly.
Another approach presented in~\cite{Kaden2019}, classified as a two-step approach, combines sampling-based planning, i.e., Rapidly Exploring Random Trees (RRTs), see, e.g.,~\cite{Lavalle2000}, with Gaussian Mixture Models and STOMP, see ~\cite{Kalakrishnan2011}, for trajectory smoothing. Note that in \cite{Kaden2019} the state costs take into account the corresponding manipulability. 
The main advantage of this approach is the capability to escape local minima with sampling-based planning compared to directly optimizing the trajectory. 
The two-step process, on the other hand, is computationally expensive.

\textcolor{red}{The main contribution of this work is a novel singularity avoidance concept based on known singularities and potential functions.}
The proposed approach is compared to manipulability maximization in terms of trajectory quality in the vicinity of singularities and suitability for online trajectory optimization. 
For a representative evaluation, three serial manipulators are considered in this work. The 6-DoF industrial \comau represents a commonly used industrial robot. As examples of collaborative robots, the non-offset 7-DoF \kuka and the 7-DoF offset \panda are investigated. Due to the offset in the kinematics, the singular configurations of the \panda differ significantly 
from the \kuka. 
To the best of the authors' knowledge, the analytic singular configurations of the \comau and the \panda have not been presented in the literature so far. 

The paper is organized as follows:
In Section~\ref{section: Mathematical Model}, the kinematic model of a rigid-body serial manipulator is described and the singular configurations are determined. In addition, the trajectory optimization used in this work is introduced.
The novel singularity avoidance approach for trajectory optimization based on potential functions is presented in Section~\ref{section: singularity avoidance}.
The statistical evaluation of the proposed approach and the manipulability maximization is discussed in Section~\ref{section: results}. Finally, Section~\ref{section: Conclusion} concludes the paper and gives an outlook on future work.

\section{Mathematical Model}
\label{section: Mathematical Model}
In this section, the derivation of the forward kinematics and the manipulator Jacobian of serial manipulators are presented. Furthermore, the dynamics of the rigid-body manipulators are described including a Cartesian inverse dynamics control law. The compensation of the nonlinear dynamics justifies the use of a linear model for the presented trajectory optimization approach.

\subsection{Forward kinematics}
In robotics, the forward kinematics determines the geometric relation between the joint space, i.e., coordinates of the robot joints, and the operational space, i.e., the position in 3D space and the orientation of the robot end-effector. The geometric relation can be systematically computed by using the homogeneous transformation
\begin{equation}
    \textcolor{red}{\mathbf{H}_{n}^{m}} = \begin{bmatrix}
        \mathbf{R}_n^m & \mathbf{d}_n^m \\
        \mathbf{0} & 1
    \end{bmatrix} \Comma
\end{equation}
where the distance vector $\mathbf{d}_n^m \in \mathbb{R}^3$ and the orthogonal rotation matrix $\mathbf{R}_n^m \in SO(3)$ represent the translation of the origin and the rotation from the coordinate frame $n$ to the coordinate frame $m$, respectively. Pure translations the in direction of the local axis $i \in \{x,y,z\}$ by the length $d$ and pure rotations by the angle $\varphi$ around the local axis $i$ are denoted by $\mathbf{H}_{\mathrm{T}i,d}$ and $\mathbf{H}_{\mathrm{R}i,\varphi}$, respectively. 

\begin{figure}
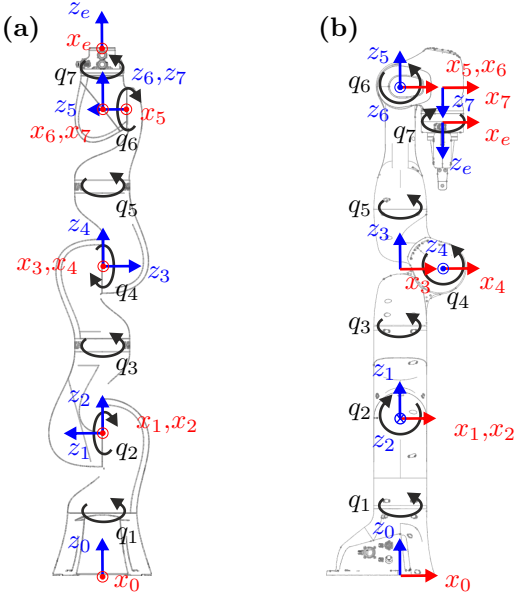

        \centering
        \quad
        \begin{subfigure}[t]{0.4\columnwidth}
            \scalebox{1.4}{\input{./figure/iiwa.tex}}
        \end{subfigure}
        \begin{subfigure}[t]{0.4\columnwidth}
            \scalebox{1.4}{\input{./figure/panda.tex}}
        \end{subfigure}
        \caption{(a) Schematics of the \kuka on the left-hand side, and (b) schematics of the \panda on the right-hand side.}
        \label{fig: KUKA_PANDA schematic}
\end{figure}

Homogeneous transformations of serial robot manipulators are described using successive homogeneous transformations $\mathbf{H}_{i-1}^i$ with the Denavit-Hartenberg (DH) convention, see \cite{spong2006robot}, consisting of two rotations and two translations in the form
\begin{equation}
    \mathbf{H}_{i-1}^i = \mathbf{H}_{\mathrm{R}z,\theta_i} \mathbf{H}_{\mathrm{T}z,d_i}\mathbf{H}_{\mathrm{T}x,a_i}\mathbf{H}_{\mathrm{R}x,\alpha_i}\Comma
\end{equation}
with the four parameters $d_i$, $\theta_i$, $a_i$, and $\alpha_i$ for each joint $i$.
Thus, the forward kinematics of a rigid-body manipulator with $M$ joints is computed in the form
\begin{equation}
    \mathbf{H}_0^M = \mathbf{H}_0^1 \mathbf{H}_1^2 ... \mathbf{H}_{M-1}^M
\Comma
\label{eq: FK general}        
\end{equation}
The \comau consists of 6 rotational joints. Its schematics and DH parameters are listed in Fig. \ref{fig: Comau schematic} and Tab. \ref{table: DH parameters}(a), respectively. The \kuka is a non-offset 7-DoF serial manipulator with all offset parameters $a_i$ equal to zero, see Tab. \ref{table: DH parameters}(b) and Fig. \ref{fig: KUKA_PANDA schematic}(a) and the \panda is an offset 7-DoF serial manipulator with non-zero parameters $a_i$, see Tab. \ref{table: DH parameters}(c) and Fig. \ref{fig: KUKA_PANDA schematic}(b). 

\begin{figure}
    \centering
    \scalebox{1.2}{\input{./figure/comau.tex}}
    \caption{Schematic of the \comau.}
    \label{fig: Comau schematic}
\end{figure}
%
%
\begin{table}[h]
        \centering
    \captionsetup{width=0.5\textwidth}        
	\caption{Denavit-Hartenberg parameters.}
        \textbf(a) \comau
	\begin{tabular}{c|cccc||}
		$i$ & $\theta_i$ & $d_i$  & $a_i$  & $\alpha_i$  \\
		\hline
		$1$ & $q_1$ & $d_1$ & $a_1$ & $\frac{\pi}{2}$\\
		$2$ & $q_2$ & $0$ & $a_2$ & $0$\\
		$3$ & $q_3$ & $0$ & $a_3$ & $\frac{\pi}{2}$\\
	\end{tabular}%
	\begin{tabular}{c|cccc}
		$i$ & $\theta_i$ & $d_i$  & $a_i$  & $\alpha_i$  \\
		\hline
		$4$ & $q_4$ &  $d_4$ & $0$ & $-\frac{\pi}{2}$\\
		$5$ & $q_5$ & $0$ & $0$ & $\frac{\pi}{2}$\\
		$6$ & $q_6$ & $d_6$ & $0$ & $0$\\
            
	\end{tabular}%

        \vspace{.5cm}
	\textbf(b) \kuka
	  \begin{tabular}{c|cccc||}
		$i$ & $\theta_i$ & $d_i$  & $a_i$  & $\alpha_i$  \\
		\hline
		$1$ & $q_1$ & $d_1$ & 0 & $\frac{\pi}{2}$\\
		$2$ & $q_2$ & $0$ & $0$ &  $-\frac{\pi}{2}$\\
		$3$ & $q_3$ & $d_3$ & $0$ & $-\frac{\pi}{2}$\\
            $4$ & $q_4$ & $0$ & $0$ & $\frac{\pi}{2}$\\
	\end{tabular}%
	\begin{tabular}{c|cccc}
		$i$ & $\theta_i$ & $d_i$  & $a_i$  & $\alpha_i$  \\
		\hline
		$5$ & $q_5$ &  $d_5$ & $0$ & $\frac{\pi}{2}$\\
		$6$ & $q_6$ & $0$ & $0$ & $-\frac{\pi}{2}$\\
		$7$ & $q_7$ & $0$ & $0$ & $0$\\
            $e$ & $0$ & $d_\mathrm{e}$ & $0$ & $0$\\
	\end{tabular}%

        \vspace{.5cm}
        \textbf(c) \panda
        \begin{tabular}{c|cccc||}
		$i$ & $\theta_i$ & $d_i$  & $a_i$  & $\alpha_i$  \\
		\hline
		$1$ & $q_1$ & $d_1$ & 0 & $0$\\
		$2$ & $q_2$ & $0$ & $0$ &  $-\frac{\pi}{2}$\\
		$3$ & $q_3$ & $d_3$ & $0$ & $\frac{\pi}{2}$\\
            $4$ & $q_4$ & $0$ & $a_4$ & $\frac{\pi}{2}$\\
	\end{tabular}%
	\begin{tabular}{c|cccc}
		$i$ & $\theta_i$ & $d_i$  & $a_i$  & $\alpha_i$  \\
		\hline
		$5$ & $q_5$ &  $d_5$ & $a_5$ & $-\frac{\pi}{2}$\\
		$6$ & $q_6$ & $0$ & $0$ & $\frac{\pi}{2}$\\
		$7$ & $q_7$ & $0$ & $a_7$ & $\frac{\pi}{2}$\\
            $e$ & $0$ & $d_\mathrm{e}$ & $0$ & $0$\\
	\end{tabular}%
 \label{table: DH parameters}
\end{table}

Substituting the DH parameters of an $M$-DoF serial manipulator in Tab. \ref{table: DH parameters} into (\ref{eq: FK general}) with the joint coordinates $\mathbf{q}^\mathrm{T} = [q_1,q_2,...,q_M]$ yields the homogeneous transformation of the Cartesian end-effector pose 
\begin{equation}
    \mathbf{H}_\mathrm{e}(\mathbf{q})    = 
    \mathbf{H}_0^M = 
    \begin{bmatrix}
        \vect{R}_\mathrm{e}(\vect{q}) & \vect{p}_\mathrm{e}(\vect{q}) \\
        \vect{0} & 1
        \end{bmatrix}
        \Comma
\label{eq: end-effector pose}        
\end{equation}
where $\mathbf{p}_\mathrm{e}$ and $\mathbf{R}_\mathrm{e}$ are the end-effector position and orientation expressed in the world frame, respectively.

\subsection{Manipulator Jacobian}
The velocity of the end-effector pose (\ref{eq: end-effector pose}) is computed by using the geometric manipulator Jacobian $\vect{J}_\mathrm{e}(\vect{q})$ as
\begin{align}
  \begin{bmatrix}
    \dot{\vect{p}}_\mathrm{e} \\
    \vect{\omega}_\mathrm{e}
  \end{bmatrix} =
  \begin{bmatrix}
    \vect{J}_\mathrm{e,v}(\vect{q}) \\
    \vect{J}_{\mathrm{e},\omega}(\vect{q})
  \end{bmatrix} \dot{\vect{q}} = \vect{J}_\mathrm{e}(\vect{q})\dot{\vect{q}}\Comma
\label{eq: jacobian relationship}  
\end{align}
with the linear end-effector velocity $\dot{\vect{p}}_\mathrm{e}$ and the angular end-effector velocity $\vect{\omega}_\mathrm{e}^\mathrm{T}=[\omega_x,\omega_y,\omega_z]$. The latter results from the skew-symmetric matrix operator $\mathbf{S}(\bm{\omega}_\mathrm{e})$ as
\begin{equation}
    \mathbf{S}(\bm{\omega}) = \dot{\mathbf{R}}_\mathrm{e}{\mathbf{R}}_\mathrm{e}^\mathrm{T} = \begin{bmatrix}
        0 & -\omega_z & \omega_y \\
        \omega_z & 0 &-\omega_x \\
        -\omega_y & \omega_x & 0
    \end{bmatrix}.
\end{equation}
In other words, the geometric manipulator Jacobian $\mathbf{J}_\mathrm{e}(\mathbf{q})$ describes the relationship between the joint-space velocity $\dot{\mathbf{q}} \in \mathbb{R}^{M}$ and the translational and angular velocities in the task space. 
\subsection{Singularity Analysis}
\label{sec: singularity analysis}
A robot configuration, causing the robot end-effector to lose the ability to move in one or more directions is called a singularity. 
Using (\ref{eq: jacobian relationship}), the task-space end-effector velocity $\mathbf{v}_\mathrm{ts}^\mathrm{T}=[\dot{\mathbf{p}}^\mathrm{T}_\mathrm{e},\bm{\omega}^\mathrm{T}_\mathrm{e}]$ is expressed in the form
\begin{equation}
\begin{aligned}
    \mathbf{v}_\mathrm{ts}=
    \begin{bmatrix}
        \dot{\mathbf{p}}_\mathrm{e} \\
        \bm{\omega}_\mathrm{e}
    \end{bmatrix}
    &=
    \begin{bmatrix}
        \mathbf{J}_{\mathrm{e},c_1} & \cdots  &\mathbf{J}_{\mathrm{e},c_M}
    \end{bmatrix}
    \begin{bmatrix}
        \dot{q}_1 \\
        \vdots \\
        \dot{q}_M
    \end{bmatrix} \\
    &= \sum_{i=1}^M \mathbf{J}_{\mathrm{e},c_i} \dot{q}_i \Comma
\end{aligned}    
\end{equation}
where $\mathbf{J}_{\mathrm{e},c_i}$ is the $i$-th column of the Jacobian $\mathbf{J}_\mathrm{e}(\mathbf{q})$. 
Hence, the geometric manipulator Jacobian $\vect{J}_\mathrm{e}(\vect{q})$ can be utilized to identify singular configurations. A robot configuration $\mathbf{q}$ is singular if 
\begin{equation}
    \mathrm{Rank} (\mathbf{J}_\mathrm{e}(\mathbf{q})) < 6
    \Comma
    \label{eq: rank Je}
\end{equation}
\textcolor{red}{since the considered workspace of the robot consists of six degrees of freedom in Cartesian space}
or, similarly,
\begin{equation}
    m(\mathbf{q}) = \sqrt{\mathrm{det}\big(\mathbf{J}_\mathrm{e} \mathbf{J}_\mathrm{e}^\mathrm{T}\big)} = 0 \Comma
    \label{eq: m(q)}
\end{equation}
which is called singularity index $m(\mathbf{q})$, see \cite{Yoshikawa1985}. Note that the mathematical definition of a kinematic singularity, e.g., (\ref{eq: rank Je}), is independent of the choice of the reference frame of the end-effector Jacobian $\mathbf{J}_\mathrm{e}(\mathbf{q})$, see, \cite{lynch2017modern}. To reduce the complexity of the expression of $\mathbf{J}_\mathrm{e}(\mathbf{q})$ in the world frame, the manipulator Jacobian is transformed into the end-effector frame \textcolor{red}{$M$}
\begin{equation}
    \mathbf{J}_\mathrm{e}^M(\mathbf{q}) = \begin{bmatrix}
        \mathbf{R}_\mathrm{e}^\mathrm{T}\textcolor{red}{(\mathbf{q})} & \mathbf{0} \\
        \mathbf{0} & \mathbf{R}_\mathrm{e}^\mathrm{T}\textcolor{red}{(\mathbf{q})} \\
    \end{bmatrix}    
    \mathbf{J}_\mathrm{e}(\mathbf{q}) \Comma
\end{equation}
for the singularity analysis. 

In the following, the general scheme for singularity analysis of the \comau is presented. Based on this scheme, the singularity analysis of the \kuka and the \panda are briefly summarized. 
The symbolic expression of the manipulator Jacobian of the 6-DoF \comau in its end-effector frame is of the form
\begin{equation}
\mathbf{J}_\mathrm{e}^6(\mathbf{q}) = 
\begin{bmatrix}
    \mathbf{J}_{\mathrm{e},r_1}^6 \\
    \vdots \\
    \mathbf{J}_{\mathrm{e},r_6}^6 
\end{bmatrix}
=
\mleft[
    \begin{array}{c c c | c c c}
        &  & & 0 & d_6 & 0 \\
        & \mathbf{J}_{\mathrm{e},11}^6 & & \sin(q_5)d_6 & 0 & 0 \\
        &  & & 0 & 0 & 0 \\
        \hline
        &  & & & & \\
        & \mathbf{J}_{\mathrm{e},21}^6 & & & \textcolor{red}{\mathbf{J}_{\mathrm{e},22}^6} & \\
        &  & & & & \\
    \end{array}
\mright] \Comma
    \label{eq: J_e_6}
\end{equation}
where $\mathbf{J}_{\mathrm{e},r_i}^6$ is the $i$-th row of $\mathbf{J}_{\mathrm{e}}^6(\mathbf{q})$, $\mathbf{J}_{\mathrm{e},11}^6 \in \mathbb{R}^{3 \times 3}$, $\mathbf{J}_{\mathrm{e},21}^6 \in \mathbb{R}^{3 \times 3}$ and
\begin{equation}
    \vect{J}^6_{\mathrm{e}, 22} = \begin{bmatrix}
        -\sin(q_5) & 0 & 0 \\
        0 & 1 & 0 \\
        \cos(q_5) & 0 & 1
    \end{bmatrix}\Comma
\end{equation}
are submatrices of $\mathbf{J}_\mathrm{e}^6(\mathbf{q})$. Applying the following row operations, see, e.g., \cite{xu2015singularity}, 
\begin{equation*}
    \begin{aligned}
    (\mathrm{A})\:\: & \mathbf{J}_{\mathrm{e},r_1} - d_6 \mathbf{J}_{\mathrm{e},r_5} \rightarrow  \mathbf{J}_{\mathrm{e},r_1} \\
    (\mathrm{B})\:\: & \mathbf{J}_{\mathrm{e},r_2} + d_6 \mathbf{J}_{\mathrm{e},r_4} \rightarrow  \mathbf{J}_{\mathrm{e},r_2} 
    \end{aligned}
    \Comma
\end{equation*}
to $\mathbf{J}_\mathrm{e}^6$ from (\ref{eq: J_e_6}) results in
\begin{equation}
\mathbf{J}_\mathrm{e}^{6\mathrm{t}} = 
\mleft[
        \begin{array}{c c c | c c c}
        &  & & 0 & 0 & 0 \\
        & \mathbf{J}_{\mathrm{e},11}^{6\mathrm{t}} & & 0 & 0 & 0 \\
        &  & & 0 & 0 & 0 \\
        \hline
        &  & &  &  &  \\
        & \mathbf{J}_{\mathrm{e},21}^{6} & &  & \mathbf{J}_{\mathrm{e},22}^{6} &  \\
        &  & &  &  &  \\
    \end{array}
\mright]
\FullStop
\label{eq: isolation rank 1}
\end{equation}
\textcolor{red}{The superscript $t$ is added to denote transformed matrices.}
Note that the rank of a matrix remains invariant under elementary row operations. 
Since $\mathbf{J}_{\mathrm{e},11}^{6\mathrm{t}}$ and $\mathbf{J}_{\mathrm{e},22}^{6}$ have to be full rank for $\mathbf{J}^6_{\mathrm{e}}$ to be non-singular,
the determinants of $\mathbf{J}_{\mathrm{e},11}^{6\mathrm{t}}$ and $\mathbf{J}_{\mathrm{e},22}^{6}$ are computed to identify the singular configurations
\begin{subequations}
    \label{eq: Comau singularity confs}
    \begin{align}
    (\mathrm{A})\:\: & q_5 = 0 \\
    (\mathrm{B})\:\: & q_3 = \arctan\left(\dfrac{d_4}{a_3}\right) \\
    \label{eq: Comau sing eqn}
    (\mathrm{C})\:\: & a_1+\cos(q_2)a_2+a_3 \cos(q_2+q_3)+
    \\&d_4\sin(q_2+q_3) = 0 \FullStop \nonumber
    \end{align}
    \label{eq: Comau sing conds}
\end{subequations}
In a similar way, the singularity analysis of a 7-DoF serial manipulator is performed by applying row operations of the corresponding manipulator Jacobian in the end-effector coordinate $\mathbf{J}_\mathrm{e}^7$. This leads to the transformed manipulator Jacobian $\mathbf{J}_\mathrm{e}^{7\mathrm{t}}$ consisting of four submatrices in the form
\begin{equation}
    \mathbf{J}_\mathrm{e}^{7\mathrm{t}} = 
    \mleft[
            \begin{array}{c c c | c c c}
            &  & & 0 & 0 & 0 \\
            & \mathbf{J}_{\mathrm{e},11}^{7\mathrm{t}} & & 0 & 0 & 0 \\
            &  & & 0 & 0 & 0 \\
            \hline
            &  & &  &  &  \\
            & \mathbf{J}_{\mathrm{e},21}^{7\mathrm{t}} & &  & \mathbf{J}_{\mathrm{e},22}^{7\mathrm{t}} &  \\
            &  & &  &  &  \\
        \end{array}
    \mright]
    \Comma    
    \label{eq: 7-DoF isolated rank}
\end{equation}
where $\mathbf{J}_{\mathrm{e},11}^{7\mathrm{t}} \in \mathbb{R}^{3 \times 4}$ and $ \mathbf{J}_{\mathrm{e},22}^{7\mathrm{t}} \in \mathbb{R}^{3 \times 3}$. 
Here, similar to (\ref{eq: isolation rank 1}), $\mathbf{J}_{\mathrm{e},11}^{7\mathrm{t}}$ must have full rank. 
Since $\mathbf{J}_{\mathrm{e},11}^{7\mathrm{t}} \in \mathbb{R}^{3 \times 4}$ is a non-square matrix, the determinant
\begin{equation}
    \mathrm{det}(\mathbf{J}_{\mathrm{e},11}^{7\mathrm{t}} \big(\mathbf{J}_{\mathrm{e},11}^{7\mathrm{t}}\big)^\mathrm{T})
    \label{eq: isolated rank det}
\end{equation}
is checked instead.
To reduce the computational complexity of (\ref{eq: isolated rank det}), the Cauchy-Binet theorem is used, see, e.g., \cite{knill2014cauchy}. This allows to express (\ref{eq: isolated rank det}) as a sum of squares in the form
\begin{equation}
   \mathrm{det}(\mathbf{J}_{\mathrm{e},11}^{7\mathrm{t}} \big(\mathbf{J}_{\mathrm{e},11}^{7\mathrm{t}}\big)^\mathrm{T}) = \sum_{s\in S}{\left[\mathrm{det}(\mathbf{J}^{7\mathrm{t}}_{\mathrm{e},11,s})\right]^2} \Comma
    \label{eq: SOS}
\end{equation}
where $S$ is the set of $3$-combinations of $\{1,...,4\}$ and $\mathbf{J}^{7\mathrm{t}}_{\mathrm{e},11,s}\in \mathbb{R}^{3 \times 3}$ is formed by columns of $\mathbf{J}^{7\mathrm{t}}_{\mathrm{e},11}$ at indices from the subset $s$. Note that a $3$-combination of $\{1,...,4\}$ is a subset of three distinct elements of the set $\{1,...,4\}$, see \cite{roberts2009applied}.
Thereby, singular configurations can be identified by analyzing the determinants of $\mathbf{J}^{7\mathrm{t}}_{\mathrm{e},11,s}$, which is much simpler than considering (\ref{eq: isolated rank det}). 
On the other hand, if $\mathbf{J}_{\mathrm{e},22}^{7\mathrm{t}}$ is rank deficient it is only a necessary condition for $\mathbf{J}_{\mathrm{e}}^{7\mathrm{t}}(\mathbf{J}_{\mathrm{e}}^{7\mathrm{t}})^\mathrm{T}$ to become singular.
Therefore, $\mathrm{det}(\mathbf{J}_{\mathrm{e}}^{7\mathrm{t}}(\mathbf{J}_{\mathrm{e}}^{7\mathrm{t}})^\mathrm{T})$ has to be checked again at the singular configurations obtained from solving $\mathrm{det}(\mathbf{J}_{\mathrm{e},22}^{7\mathrm{t}}) = 0$.
Due to the limited length of the paper, the detailed analysis of (\ref{eq: SOS}) for the \kuka and the \panda is omitted here and all singular configurations for these 7-DoF robots are listed in the Appendix.
 
\subsection{Dynamics}

The dynamical rigid-body model of a serial manipulator is given by
\begin{align}
 \vect{M}(\vect{q})\ddot{\vect{q}} + \vect{C}(\vect{q}, \dot{\vect{q}})\dot{\vect{q}} + \vect{g}(\vect{q}) = \vect{\tau}\Comma
\end{align}
with the positive definite inertia matrix $\vect{M}(\vect{q})$, the Coriolis matrix $\vect{C}(\vect{q}, \dot{\mathbf{q}})$ and the vector of gravitational forces $\vect{g}(\vect{q})$. The generalized torques are denoted by $\vect{\tau}$.
It is assumed that the nonlinear dynamics are compensated by an inverse dynamics control law of the form
\begin{align}
  \label{eqn:computed_torque}
  \vect{\tau} = \vect{M}(\vect{q})\vect{v} + \vect{C}(\vect{q}, \dot{\vect{q}})\dot{\vect{q}} + \vect{g}(\vect{q})\Comma
\end{align}
resulting in the remaining linear dynamics
\begin{align}
  \label{eqn:double_integrator}
  \ddot{\vect{q}} = \vect{v}\Comma
\end{align}
with the virtual input $\vect{v}$. Different approaches exist for stabilizing the linear system including Cartesian impedance control. In the following, a reference trajectory is planned for this remaining linear dynamics.

\subsection{Trajectory Optimization}
\label{sec: trajectory optimization}

The trajectory optimization problem is formulated in the joint space.
If Cartesian reference trajectories are required, they can be uniquely calculated using the forward kinematics and differential kinematics of the manipulator. 
The trajectory optimization is implemented using a direct transcription method, discretizing the trajectory into $N$ grid points and solving the discrete optimization problem
\begin{subequations}
  \begin{alignat}{2}
    &\min_{\bm{\xi}} && t_\mathrm{F}+\sum_{k = 0}^{N - 1} \big[\vect{v}^\mathrm{T}_k \vect{v}_k + L_{\mathrm{sing}}(\vect{q}_k)\big]
    \label{eqn: traj_opt}\\
    &\quad \text{s.t.} && \quad \vect{x}_{k + 1} = \vect{\Phi} \vect{x}_{k } + \vect{\Gamma} \vect{v}_{k} \label{eqn:opt_dyn}\\
    & \quad && \quad \vect{x}_{0} = \vect{x}_{\mathrm{S}} \:\:,\:\: \vect{x}_{N - 1} = \vect{x}_{\mathrm{T}} \label{eqn:init_cond_1_2}  \\
    &                  \quad && \quad \underline{\vect{x}} \le \vect{x}_{k} \le \overline{\vect{x}}, \quad k = 0, \dots, N - 1 \label{eqn:x_limit} \\
    &             \quad && \quad \underline{\vect{v}} \le \vect{v}_{k} \le \overline{\vect{v}}, \quad k = 0, \dots, N - 1 \label{eqn:u_limit} \\
    & \quad && \quad t_\mathrm{F} \ge 0 \label{eqn:t_f_limit}
  \end{alignat}
  \label{eqn:traj_opt}
\end{subequations}
with the state $\vect{x}^\mathrm{T}_k = [\vect{q}_k^\mathrm{{T}}, \dot{\vect{q}}_k^\mathrm{T} ]$, the input $\vect{v}_k$, and 
\begin{equation*}
    \bm{\xi}^\mathrm{T} = [t_\mathrm{F},\vec{x}_{0}, \dots, \vect{x}_{N - 1},\\ \vect{v}_{0}, \dots, \vect{v}_{N - 1}]
    \FullStop
\end{equation*}
The constraints (\ref{eqn:init_cond_1_2}) describe the start- and the end-point constraints. The constraints on the optimization variables are given in (\ref{eqn:x_limit}), (\ref{eqn:u_limit}), and (\ref{eqn:t_f_limit}). The discretized dynamics of the double integrator system (\ref{eqn:double_integrator}) with the sampling time $h = t_\mathrm{F} / N$ reads as
\begin{subequations}
    \begin{align}
  \vect{\Phi} &= \begin{bmatrix}
                  1 & h \\
                  0 & 1
                \end{bmatrix}\otimes \vect{I}_{M}
  \\              
  \vect{\Gamma} &= \begin{bmatrix}
                  \frac{h^2}{2} \\
                  h
               \end{bmatrix} \otimes \vect{I}_{M}\FullStop\label{eqn:sys_d_mat}
\end{align}
\end{subequations}

The operator $\otimes$ describes the Kronecker product and $\vect{I}_M$ is the identity matrix of size $M$.
The objective function consists of minimizing the time $t_\mathrm{F}$, a regularization term $\vect{v}_k^\mathrm{T}\vect{v}_k$, and a singularity avoidance term $L_{\mathrm{sing}}(\vect{q}_k)$.
The objective functions for singularity avoidance considered in this paper are discussed in the following section.
\textcolor{red}{The optimization problem is solved using the interior-point solver IPOPT, see~\cite{Waechter2006}, discussed in Section~\ref{section: results}.}

\section{Singularity Avoidance}
\label{section: singularity avoidance}
This section presents two objective function formulations for singularity avoidance. 
First, an approach for directly maximizing the manipulability known from the literature is discussed. 
Second, a novel approach formulating the singularity avoidance using potential functions for known singularities based on the calculations in Section~\ref{sec: singularity analysis} is presented.

Strict singularity avoidance can be achieved by constraining the kinematic manipulability measure (\ref{eq: m(q)}) to stay above a certain minimum value, as done, e.g., in~\cite{Marani2002}. 
For fast online planning, however, a formulation based on an objective function is often preferable and leads to a faster convergence of the optimization. 
To maximize manipulability, the objective function can be formulated as
\begin{equation}
    \label{eq: L_sing_1}
    L_{\mathrm{sing}, 1}(\vect{q}) = \frac{w_m}{m(\vect{q}) + \varepsilon} \FullStop
\end{equation}
The parameter $\varepsilon > 0$ ensures that the expression remains well defined even for $m(\vect{q}) = 0$ and $w_m > 0$ is a weighting parameter.
Note that this does not strictly avoid singularities in all cases but leads to good results in practise provided that the weight $w_m$ is chosen large enough compared to the other terms in the objective function.
The main disadvantage of this formulation is that the evaluation of $m(\vect{q})$, see, e.g., (\ref{eq: m(q) COMAU}) and (\ref{eq: m(q) KUKA}), is computationally expensive especially for the gradients required for the optimization algorithm.
Hence, this approach is not suitable for fast optimization-based planning. In addition, since the manipulability measure is related only to the volume of the manipulability ellipsoid, it does not necessarily indicate a short distance to a singularity.

To this end, a different approach based on potential functions is proposed, where the objective function is increased close to singular configurations of the manipulator. 
Common structures of serial manipulators allow the calculation of those singular configurations analytically via the manipulator Jacobian (\ref{eq: rank Je}) in form of explicit or implicit equations, see, e.g., (\ref{eq: Comau singularity confs}) or (\ref{eq: PANDA sing conditions}). These equations are less complex compared to the manipulability measure~(\ref{eq: m(q)}).
From this calculation, implicit functions of the form $\psi(\vect{q}) = 0$ can be obtained that are exactly zero if the manipulator Jacobian is singular.

For example, the singularities of the \kuka are caused by one or two joints, see (\ref{eq: KUKA sing conditions}). 
Hence, for a single-joint singularity condition, the $i$-th equation reads as
\begin{align}
    \label{eq: iiwa_single_joint_sing}
    \psi_i(q_m) = (q_m - q_{m, s})^2
\end{align}
for a singularity at $q_{m, s}$ and
\begin{align}
    \label{eq: iiwa_double_joint_sing}
    \psi_{i}(q_m, q_n) = (q_m - q_{m, s})^2 + (q_n - q_{n, s})^2 
\end{align}
for a singularity caused by two joints $q_{m}$ and $q_{n}$ at $q_{m, s}$ and $q_{n, s}$ simultaneously.
\textcolor{red}{}
Given a general function $\psi_i(\vect{q})$, the potential function for a singularity is then introduced in the form
\begin{align}
  \varphi_i(\vect{q}) = \exp\left(\eta_{1} - \eta_{2} \psi_i(\vect{q}) \right) \FullStop
\end{align}
Note that the parameters $\eta_{1} > 0$ and $\eta_{2} > 0$ define the peak value and the width of the potential function, respectively.

The overall cost function for singularity avoidance $L_{\mathrm{sing}, 2}(\vect{q})$ is then given by
\begin{align}
  L_{\mathrm{sing}, 2}(\vect{q}) &= w_m\sum_{i} \varphi_i(\vect{q})
  \FullStop
  \label{eq: proposed SA}
\end{align}
\textcolor{red}{The individual potential functions are equally important for singularity avoidance and are therefore equally weighted in the sum.}
The gradient and Hessian of~(\ref{eq: proposed SA}) can be easily calculated and the computational costs are significantly smaller compared to~(\ref{eq: m(q)}).
Another advantage of this approach is that (\ref{eq: iiwa_single_joint_sing}) and (\ref{eq: iiwa_double_joint_sing}) are joint-space distance measures directly for the singularities instead of the volume proxy utilized in the singularity index~(\ref{eq: m(q)}).


\section{Results}
\label{section: results}
In this section, the proposed singularity avoidance concept for online trajectory optimization is evaluated in comparison to the direct manipulability maximization.
For this, Monte Carlo simulations are performed with the three robots from Section~\ref{section: Mathematical Model}. For each robot, $N_{\mathrm{mc}}=10^4$ pairs of initial robot configurations $\mathbf{x}_\mathrm{S}^\mathrm{T} = [\mathbf{q}^\mathrm{T}_0,\mathbf{0}^\mathrm{T}]$ and target configurations $\mathbf{x}_\mathrm{T}^\mathrm{T} = [\mathbf{q}^\mathrm{T}_\mathrm{T},\mathbf{0}^\mathrm{T}]$ are randomly selected in the workspace from a uniform distribution in the admissible ranges. In addition, configurations with a manipulability of less than $\num{1e-4}$ are excluded.
\textcolor{red}{The parameters for the trajectory optimization are summarized in Tab.~\ref{tab: parameters}.}
\begin{table}
    \captionsetup{width=0.5\textwidth}
    \caption{Parameters for the singularity avoidance cost functions $L_{\mathrm{sing}, 1}$ and $L_{\mathrm{sing}, 2}$.} 
    \label{tab: parameters}
    \begin{center}
        \begin{threeparttable}
        \begin{tabular}{c c}
             parameter & value \\ 
            \hline
            $w_m$ & $10^2$\\  
            $\varepsilon$ & $10^{-6}$ \\
            $\eta_{1}$ & $\log(2400)$ \\
            $\eta_{2}$ & $400$ \\
            \hline
        \end{tabular}
        \end{threeparttable}
    \end{center}
\end{table}
The trajectory optimization (\ref{eqn:traj_opt}) is evaluated for three different objectives, i.e. without singularity avoidance, with the manipulability-based objective (\ref{eq: L_sing_1}), and with the proposed singularity avoidance approach (\ref{eq: proposed SA}). The simulation results are obtained on a PC with $3.4$ GHz Intel Core i7 and $32$ GB of RAM. The interior-point solver IPOPT, see~\cite{Waechter2006}, with the linear solver MA27, see \cite{duff2004ma57}, is employed to solve the optimization problem~(\ref{eqn:traj_opt}). Additionally, the gradient and the Hessian are computed using CasADi, see~\cite{Andersson2019}. The trajectory optimization in~(\ref{eqn:traj_opt}) is discretized with $N=30$ collocation points giving a total of $631$ optimization variables for the $7$ DoF and $541$ for the $6$ DoF manipulator. 

To compare the performance of the three objective functions, the following indices are computed and are evaluated statistically. 
\begin{itemize}
    \item The minimum value of the manipulability measure (\ref{eq: m(q)}) over all Monte Carlo samples 
    \begin{equation}
        \begin{aligned}
            &m_{\mathrm{min}} = \min m(\vect{q}_{k, i}) \Comma \\
            &k = 0,\dots, N - 1, \:\: i = 1, \dots, N_{\mathrm{mc}} \FullStop
        \end{aligned}
        \label{eq: m_min}
    \end{equation}
    \item The maximum value of the manipulability measure (\ref{eq: m(q)}) over all Monte Carlo samples
    \begin{equation}
        \begin{aligned}
            &m_{\mathrm{max}} = \max m(\mathbf{q}_{k,i}),\:  \\
            &k = 0,\dots, N - 1, \:\: i = 1,\dots, N_{\mathrm{mc}} \FullStop
        \end{aligned}
        \label{eq: m_max}
    \end{equation}
    \item The average value of the manipulability measure (\ref{eq: m(q)}) for each Monte Carlo sample $i = 1, \dots, N_{\mathrm{mc}}$ 
    \begin{equation}
        \begin{aligned}
            &m_{\mathrm{avg}, i} = \dfrac{1}{N} \sum_{k = 0}^{N - 1} m(\vect{q}_{k, i}) \FullStop
        \end{aligned}
        \label{eq: m_avg}
    \end{equation}
    \item The approximate length of the 3D end-effector path for each Monte Carlo sample $i = 1, \dots, N_{\mathrm{mc}}$
    \begin{equation}
        \begin{aligned}
            &l_{\mathrm{p}, i} = \sum_{k = 1}^{N - 1}{\sqrt{\bm{\Delta}_{k, i}^\mathrm{T}\bm{\Delta}_{k, i}}} \Comma
        \end{aligned}
        \label{eq: l_p}
    \end{equation}
    with $\bm{\Delta}_{k, i} = \mathbf{p}_\mathrm{e}(\mathbf{q}_{k, i})-\mathbf{p}_\mathrm{e}(\mathbf{q}_{k-1, i})$. The function $\mathbf{p}_\mathrm{e}(\mathbf{q})$ computes the position of the end-effector in 3D, as defined in (\ref{eq: end-effector pose}).
    \item The sum of differences in rotation, see, e.g., \cite{huynh2009metrics}, along the 3D path for each Monte Carlo sample $i = 1, \dots, N_{\mathrm{mc}}$
    \begin{equation}
        \begin{aligned}
            &l_{\mathrm{quat}, i} = \sum_{k = 1}^{N - 1} {\delta}_{k, i}\Comma
        \end{aligned}
        \label{eq: l_quat}
    \end{equation}
    with ${\delta}_{k, i} = 1-\mathrm{quat}(\mathbf{R}_{\mathrm{e}}(\mathbf{q}_{k, i}))^\mathrm{T} \mathrm{quat}(\mathbf{R}_{\mathrm{e}}(\mathbf{q}_{k-1, i}))$. The function $\mathrm{quat}(\mathbf{R})$ transforms a rotation matrix $\vect{R}$ into its corresponding quaternion, see, e.g., \cite{shepperd1978quaternion}.
\end{itemize}

The results of the Monte Carlo simulation are summarized in Tab.~\ref{tab: COMAU MC}, Tab.~\ref{tab: KUKA MC}, and Tab.~\ref{tab: PANDA MC} for the \comau, the \kuka, and the \panda, respectively. For the trajectory duration $t_\mathrm{F}$, the computation time, the average manipulability of a single  trajectory (\ref{eq: m_avg}), the approximate path length (\ref{eq: l_p}) and the change in orientation~(\ref{eq: l_quat}), the mean and the standard deviation over all Monte Carlo samples are reported.
It can be observed that both singularity avoidance schemes increase the average trajectory duration compared to the baseline without singularity avoidance.
This is not surprising because avoiding a singularity typically leads to a deviation from the shortest path. 
The proposed singularity avoidance scheme, however, leads to a shorter duration of the trajectories on average compared to the manipulability maximization, except for the \comau which could be possible because less alternative trajectories are available for a non-redundant manipulator.
This can also be seen in the average path length measures (\ref{eq: l_p}) and (\ref{eq: l_quat}) being the longest for the manipulability maximization approach, also for the \comau.
The reason is that the proposed approach only influences the trajectories locally around possible singularities and not globally along the entire trajectory. 
In contrast, due to this global influence of the manipulability, the manipulability maximization approach leads to higher manipulability on average~(\ref{eq: m_avg}) along the whole trajectories. Hence, if maximization of the manipulability is desired instead of singularity avoidance, considering it directly is the better approach.
On the other hand, the minimum value of the manipulability according to~(\ref{eq: m_min}) is larger for the proposed approach, which also shows the superiority in terms of singularity avoidance.
In practice, the achieved minimum manipulability depends on the weighting of the objective terms and must be chosen carefully to avoid large control torques. 
\textcolor{red}{The maximum achievable manipulability depends on the kinematic structure of the robots. Since the workspace of the manipulators is well covered by the Monte Carlo sampling, the maximum manipulability~(\ref{eq: m_max}) achieved is very similar with both approaches.}
One of the main advantages of the proposed approach is the reduced average computation time making it more suitable for online trajectory optimization. 
For simpler kinematics, like the \comau, the difference is only a few milliseconds, for the 7 DoF \kuka a larger difference can be observed, and for the \panda with a significantly larger expression for the manipulability, this advantage is much more pronounced.
\begin{table}
    \captionsetup{width=0.5\textwidth}
    \caption{\protect\comau:\\ Performance evaluation of the proposed singularity avoidance (SA) for trajectory optimization in comparison to manipulability maximization.} 
    \label{tab: COMAU MC}
    \begin{center}
        \begin{threeparttable}
        \begin{tabular}{c c c c}
             &  without SA  & \textcolor{red}{Manip. Max.} & Proposed SA\\ 
            \hline
            $t_\mathrm{F}$ (\SI{}{\second}) & $2.3174 \pm 0.46$ &  $3.82 \pm 0.73$ &  $4.07 \pm 2.15$\\  
            \makecell{comp. \\time (\SI{}{\milli\second})} & $10.4 \pm 5.8$ &  $27.3 \pm 18.3$ &  $21.5 \pm 15.8$\\  
            $m_{\mathrm{min}}$ & $2.01^{-7}$ &  $6.6^{-6}$   &  $1.01^{-4}$\\  
            $m_{\mathrm{max}}$ & $0.466$ & $0.466$ & $0.466$ \\
            $m_{\mathrm{avg}}$ & $0.16 \pm 0.08$ & $0.362 \pm 0.04$& $0.19 \pm 0.07$\\
            $l_{\mathrm{p}}~\textcolor{red}{(\SI{}{\meter})}$ & $2.42 \pm 1.35$ &  $4.03\pm 1.6$ &  $3.86\pm 2.36$\\  
            $l_{\mathrm{quat}}$ &  $1.52 \pm 1.35$ & $2.2 \pm 1.65$ & $1.92 \pm 1.46$\\  
            \hline
        \end{tabular}
        \end{threeparttable}
    \end{center}
\end{table}
\begin{table}
    \captionsetup{width=0.5\textwidth}
    \caption{\protect\kuka:\\Performance evaluation of the proposed singularity avoidance (SA) for trajectory optimization in comparison to manipulability maximization.} 
    \label{tab: KUKA MC}
    \begin{center}
        \begin{threeparttable}
        \begin{tabular}{c c c c}
             &  without SA  & \textcolor{red}{Manip. Max.} & Proposed SA\\ 
            \hline
            $t_\mathrm{F}$ (\SI{}{\second}) & $2.67 \pm 0.48$ &  $9.75 \pm 1.73$ &  $5.03 \pm 2.59$\\  
            \makecell{comp. \\time (\SI{}{\milli\second})} & $11.4 \pm 6.3$ &  $62.3 \pm 26.7$ &  $43.8 \pm 29.1$\\  
            $m_{\mathrm{min}}$ & $5.25^{-8}$ &  $1.82^{-07}$   &  $1.007^{-4}$\\  
            $m_{\mathrm{max}}$ & $0.159$ & $0.16$ & $0.159$ \\
            $m_{\mathrm{avg}}$ & $0.061 \pm 0.027$ & $0.142 \pm 0.0035$& $0.067 \pm 0.025$\\
            $l_{\mathrm{p}}~\textcolor{red}{(\SI{}{\meter})}$ & $1.95 \pm 0.90$ &  $3.34 \pm 1.02$ &  $2.2\pm 0.94$\\  
            $l_{\mathrm{quat}}$ &  $1.65 \pm 1.43$ & $3.28 \pm 2.07$ & $1.99 \pm 1.67$\\  
            \hline
        \end{tabular}
        \end{threeparttable}
    \end{center}
\end{table}

\begin{table}
    \captionsetup{width=0.5\textwidth}
    \caption{\protect\panda:\\ Performance evaluation of the proposed singularity avoidance (SA) for trajectory optimization in comparison to manipulability maximization.} 
    \label{tab: PANDA MC}
    \begin{center}
        \begin{threeparttable}
        \begin{tabular}{c c c c}
             &  without SA  & \textcolor{red}{Manip. Max.} & Proposed SA\\ 
            \hline
            $t_\mathrm{F}$ (\SI{}{\second}) & $2.44 \pm 0.29$ &  $11.3 \pm 2.29$ &  $4.96 \pm 3.11$\\  
            \makecell{comp. \\time (\SI{}{\milli\second})} & $11.1 \pm 5.5$ &  $245.3 \pm 377$ &  $72.3 \pm 57.5$\\  
            $m_{\mathrm{min}}$ & $6.5^{-6}$ &  $6.82^{-5}$   &  $3.07^{-4}$\\  
            $m_{\mathrm{max}}$ & $0.128$ & $0.13$& $0.129$ \\
            $m_{\mathrm{avg}}$ & $0.04 \pm 0.02$ & $0.112 \pm 0.012$& $0.057 \pm 0.0188$\\
            $l_{\mathrm{p}}~\textcolor{red}{(\SI{}{\meter})}$ & $1.34 \pm 0.69$ &  $2.37\pm 0.81$ &  $1.866\pm 0.953$\\  
            $l_{\mathrm{quat}}$ &  $1.76 \pm 1.46$ & $3.62 \pm 2.13$ & $2.41 \pm 1.81$\\  
            \hline
        \end{tabular}
        \end{threeparttable}
    \end{center}
\end{table}


\section{Conclusion}
\label{section: Conclusion}
In this work, an approach for singularity avoidance based on potential functions around known singular configurations for online trajectory optimization of serial manipulators is presented. 
Singular configurations were calculated for three different robots, the \comau, the \kuka, and the \panda.
The proposed approach is compared to the well-known manipulability maximization in a Monte Carlo simulation.
The results show that the proposed approach speeds up the singularity avoidance in online trajectory optimization in all of the investigated cases. The advantage becomes more significant for more complex kinematic structures. 
On the other hand, the proposed approach is not a full replacement for manipulability maximization if high average manipulability is required.
Another benefit of the proposed approach is that shorter paths and trajectories with a shorter duration are obtained.

Future work aims at integrating the proposed singularity avoidance approach in real-time trajectory optimization for human-robot interaction applications in combination with Cartesian compliance control. Furthermore, investigating the qualitative properties of the generated trajectories with respect to their path shape is also an interesting point. 

\section*{Appendix}
\label{section: Appendix}
In the following, the abbreviations $s(\cdot) = \sin(\cdot)$ and $c(\cdot) = \cos(\cdot)$ are used for a compact notation.

\subsection*{\normalfont\comau:}
The square of the manipulability (\ref{eq: m(q)}) of the \comau reads as
\begin{equation}
\begin{aligned}
    &m^2(\mathbf{q}) = \dfrac{1}{y^2}a_2^2s^2(q_5)(a_3x-d_4)^2 \bigg[\big(-2c(q_2)a_2+a_1\big)\\
    &\big((-xd_4-a_3)c(q_2)+ s(q_2)(a_3x-d_4)\big)\sqrt{y} + \\
    &\big((a_2^2-a_3^2+d_4^2)x^2+4a_3d_4x+a_2^2+a_3^2-d_4^2\big)c^2(q_2) +\\
    &\big(-2(xd_4+a_3)(a_3x-d_4)s(q_2)+2a_1a_2y\big)c(q_2)+ \\
    &(a_1^2+a_3^2)x^2-2a_3d_4x+a_1^2+d_4^2\bigg] \Comma    
\end{aligned}
\label{eq: m(q) COMAU}
\end{equation}
with $x=\tan(q_3)$ and $y=x^2+1$. 

\subsection*{\normalfont\kuka:}

The square of the singularity index (\ref{eq: m(q)}) of the \kuka reads as
\begin{equation}
    \begin{aligned}
        &m^2(\mathbf{q}) = 2d^2_\mathrm{se}d^2_\mathrm{ew}s^2(q_4)  \\
        \bigg[&d^2_\mathrm{se}s^2(q_2)s^2(q_4)c^2(q_5)c^2(q_6)+ \\
        &d^2_\mathrm{ew}c^2(q_2)c^2(q_3)s^2(q_4)s^2(q_6)+ \\
        &\big(d^2_\mathrm{se} + 2d_\mathrm{se}d_\mathrm{ew}c(q_4)-d_\mathrm{ew}^2\big)s^2(q_2)s^2(q_6)+\\
        &\dfrac{1}{2}\big(d^2_\mathrm{se}c(q_4)+d_\mathrm{se}d_\mathrm{ew}\big)s^2(q_2)s(q_4)c(q_5)s(2q_6)+\\
        &\dfrac{1}{2}\big(d^2_\mathrm{ew}c(q_4)+d_\mathrm{se}d_\mathrm{ew}\big)s(2q_2)c(q_3)s(q_4)s^2(q_6)\bigg],
    \end{aligned}
    \label{eq: m(q) KUKA}
\end{equation}
with $d_\mathrm{se} = d_2 + d_3$ and $d_\mathrm{ew} = d_4 + d_5$. 
The singular configurations of the \kuka are given by
\begin{subequations}
    \begin{align}
        &(\mathrm{A})\:\: q_4 = 0 \Comma \label{eq: KUKA sing 1}\\
        &(\mathrm{B})\:\: q_2 = 0 \:\land\: \left(q_3 = \frac{\pi}{2} \lor q_3 = -\frac{\pi}{2}\right) \Comma \label{eq: KUKA sing 2}\\
        &(\mathrm{C})\:\: q_2 = 0 \:\land\: q_6 = 0 \Comma\\
        &(\mathrm{D})\:\: \left(q_5 = \frac{\pi}{2} \lor q_5 = -\frac{\pi}{2}\right) \:\land\: q_6 = 0\FullStop 
    \end{align}
\label{eq: KUKA sing conditions}   
\end{subequations}

\subsection*{\normalfont\panda:}
Due to the limited length of the paper, the symbolic expression of the singularity index~(\ref{eq: m(q)}) of the \panda is omitted. However, the source file is provided upon request. 
The singular configurations of the \panda are 
\begin{subequations}
    \begin{align}
        &(\mathrm{A})\:\: s(q_2) = 0 \:\land\: c(q_3)=0 \:\land\: c(q_5)=0,\\
        &(\mathrm{B})\:\: c(q_5) = 0 \:\land\: f_{\mathrm{sing},1}(q_4,q_6)=0,\\
        &(\mathrm{C})\:\: q_4 = \arctan\bigg[\dfrac{a_5(d_3+d_5)}{-a_5^2+d_5d_3}\bigg] \:\land\: s(q_5)=0,\\
        &(\mathrm{D})\:\: s(q_2) = 0 \:\land\: f_{\mathrm{sing},2}(q_3,q_4,q_5,q_6)=0\Comma
    \end{align}
\label{eq: PANDA sing conditions}      
\end{subequations}
with 
\begin{subequations}
    \begin{equation}
        \begin{aligned}
            f_{\mathrm{sing},1}&(q_4,q_6) = c(q_4)a_5[a_7+(d_3+d_5)s(q_6)]+\\
            &s(q_4)[-a_7d_3+(a_5^2-d_5d_3)s(q_6)] \Comma
        \end{aligned}
        \label{eq: eq1 panda sing}
    \end{equation}
    \begin{equation}
        \begin{aligned}
            f_{\mathrm{sing},2}&(q_3,q_4,q_5,q_6) = -a_5(x^2a_5+yxd_3+(1-y)a_5)\cdot\\
            &c(q_3)a_7c^2(q_5)+a_5s(q_3)(x^2a_5+yxd_3+(1-y)a_5)\cdot\\
            &a_7s(q_5)c(q_5)-[s(q_6)((a_5^2-d_5d_3)x+(d_3+d_5)a_5) - \\
            &a_7(d_3x-a_5)]c(q_3)(ya_5+d_5x-a_5)\Comma
        \end{aligned}
        \label{eq: eq2 panda sing}
    \end{equation}
\end{subequations}
and $x =\tan(q_4)$ and $y = \sqrt{x^2+1}$.

\bibliography{singularity_avoidance_IFAC2023}

%


\end{document}